\PassOptionsToPackage{table}{xcolor}
\documentclass[]{fairmeta}

\usepackage{amsmath,amsfonts,bm}

\def\eqref#1{equation~\ref{#1}}

\def\1{\bm{1}}

\DeclareMathAlphabet{\mathsfit}{\encodingdefault}{\sfdefault}{m}{sl}
\SetMathAlphabet{\mathsfit}{bold}{\encodingdefault}{\sfdefault}{bx}{n}

\usepackage{amsmath,amssymb,amsthm,mathtools}
\usepackage{hyperref}
\usepackage[inline]{enumitem}
\usepackage{graphicx}
\usepackage{url}
\usepackage{subfig}
\usepackage{nicefrac}
\usepackage{cleveref}
\usepackage[table]{xcolor}
\usepackage{array}
\usepackage{tabularx}
\usepackage{booktabs}
\usepackage{multirow}
\usepackage{makecell}
\usepackage{microtype}
\usepackage{float}
\usepackage{algorithm}
\usepackage{algpseudocode}
\usepackage{mdframed}
\usepackage{keyval}

\definecolor{srpoblue}{HTML}{2457A7}
\definecolor{srpohighlight}{HTML}{EAF3FF}
\definecolor{srpogreen}{HTML}{2C7A5B}
\definecolor{srpored}{HTML}{B33A3A}
\definecolor{srpogray}{HTML}{F1F3F5}
\definecolor{casegray}{HTML}{777777}
\definecolor{casepurple}{HTML}{7A657E}
\definecolor{casegold}{HTML}{A9845F}
\definecolor{casegreen}{HTML}{0B746B}
\definecolor{caseblue}{HTML}{3E658D}
\newcommand{\method}{\textsc{SRPO}}

\newtheorem{proposition}{Proposition}

\newtheorem{definition}{Definition}
\newcolumntype{L}{>{\raggedright\arraybackslash}X}
\newcolumntype{Y}{>{\centering\arraybackslash}X}

\newcommand{\seedval}[2]{\ensuremath{\mathbf{#1}\,{\scriptscriptstyle\pm\,#2}}}

\newcommand{\suppcontentsdots}{\leaders\hbox to 0.72em{\hfil.\hfil}\hfill}
\newcommand{\suppcontentssection}[2]{%
  \par\vspace{0.40em}\noindent
  \hyperref[#1]{\textcolor{srpoblue}{\Large\bfseries\makebox[2.65em][l]{\ref*{#1}}#2}}%
  \hfill\hyperref[#1]{\textcolor{black}{\Large\bfseries\pageref*{#1}}}\par\vspace{0.22em}}
\newcommand{\suppcontentssubsection}[2]{%
  \par\noindent\hspace*{2.65em}%
  \hyperref[#1]{\textcolor{srpoblue}{\large\bfseries\makebox[4.05em][l]{\ref*{#1}}#2}}%
  \nobreak\suppcontentsdots\nobreak\hyperref[#1]{\textcolor{black}{\large\bfseries\pageref*{#1}}}\par\vspace{0.10em}}
\makeatletter
\newlength{\srpocasewidth}
\define@key{srpocase}{width}{\setlength{\srpocasewidth}{#1}}
\makeatother
\newenvironment{promptbox}[1]{%
  \begin{mdframed}[
    linecolor=black,backgroundcolor=white,linewidth=.55pt,
    frametitle={#1},frametitlebackgroundcolor=black,
    frametitlefont=\color{white}\bfseries\small,
    nobreak=true,
    innerleftmargin=2.3mm,innerrightmargin=2.3mm,
    innertopmargin=1.7mm,innerbottommargin=1.7mm,
    skipabove=4pt,skipbelow=7pt]
  \small\emergencystretch=1.5em\sloppy
}{%
  \end{mdframed}%
}
\newenvironment{casebox}[3][]{%
  \setlength{\srpocasewidth}{\linewidth}%
  \setkeys{srpocase}{#1}%
  \begin{mdframed}[
    linecolor=#2,backgroundcolor=#2!3,linewidth=.55pt,
    frametitle={#3},frametitlebackgroundcolor=#2,
    frametitlefont=\color{white}\bfseries\small,
    repeatframetitle=true,
    userdefinedwidth=\srpocasewidth,align=center,
    innerleftmargin=2.5mm,innerrightmargin=2.5mm,
    innertopmargin=1.8mm,innerbottommargin=1.8mm,
    skipabove=4pt,skipbelow=5pt]
  \small\emergencystretch=1.5em\sloppy
}{%
  \end{mdframed}%
}

\title{SRPO: Setwise Relative Policy Optimization for Multi-Agent Systems}

\author[1,2]{Shengtian Yang}
\author[1]{Ziyu Xiong}
\author[1]{Yu Li}
\author[2,3]{Yewen Li}
\author[3]{Bo An}
\author[2]{Qingpeng Cai}
\author[1,\dagger]{Lei Feng}
\affiliation[1]{Southeast University, Nanjing, China}
\affiliation[2]{Kuaishou Technology, Beijing, China}
\affiliation[3]{Nanyang Technological University, Singapore}
\contribution[\dagger]{Corresponding author}

\abstract{
Multi-agent systems enable complex reasoning and tool use by coordinating agents that divide roles and refine candidate solutions. Existing methods typically update individual agent responses or treat a complete trajectory as one training example. However, these methods may produce misleading policy updates because they assign the same final outcome to responses or trajectory segments that may play different roles in different team decisions. This is because treating each response as an independent update may separate outputs that jointly determine the next action, while treating an entire trajectory as one update may combine decisions made after different observations. These limitations call for a policy update defined at the level of a team decision, outputs that lead to the same state transition are optimized under a shared objective. In this paper, we propose Setwise Relative Policy Optimization (SRPO) for multi-agent systems. We represent the outputs used together to produce one state transition as an active set. A singleton set covers division of labor, while a larger set covers joint co-evolution, and the set composition can change across decisions. SRPO assigns a shared advantage to each active set, clips the combined policy change, and normalizes its scale according to the set size. This ties each update to the decision that produced the next state. Experiments on mathematical reasoning and multi-turn search demonstrate the effectiveness of SRPO across both tasks. Additional ablation studies analyze the normalization choice and training behavior under changing active-set sizes.
}

\correspondence{\email{yangshengtian58@gmail.com}}

\begin{document}
\maketitle
\raggedbottom

\section{Introduction}

Large language models (LLMs) can use tools, retrieve evidence, and revise their answers through interaction \citep{yao2023react,schick2023toolformer,shinn2023reflexion,lyu2026agentbrew}. Multi-agent systems (MAS) bring several such agents together to solve a shared task. Their collaboration commonly takes two forms: division of labor and joint co-evolution. Division of labor assigns different roles to agents, such as proposing a solution, checking it, and producing an answer. Joint co-evolution instead lets several agents develop candidate solutions that are compared and refined together. Both forms of collaboration depend on how agents use and respond to one another's outputs. Reinforcement learning provides a way to train this collaboration using feedback on the team's final outcome \citep{lowe2017maddpg,yu2022mappo}. Existing methods use this feedback to update agent policies through role-specific training, group-relative objectives, or agent-wise advantage normalization \citep{park2025maporl,liu2025magrpo,zhao2025strongermas,chen2025mhgpo,feng2026drmas}. These updates are typically organized around individual responses, turns, or complete trajectories. Although their interaction patterns differ, both forms present the same interface to the environment: the system observes a state and returns a team action. The key difference is how many agent outputs must be collected before that action is formed, which also determines the granularity at which a policy update should be defined.

\begin{figure*}[t]
\centering
\includegraphics[width=\linewidth]{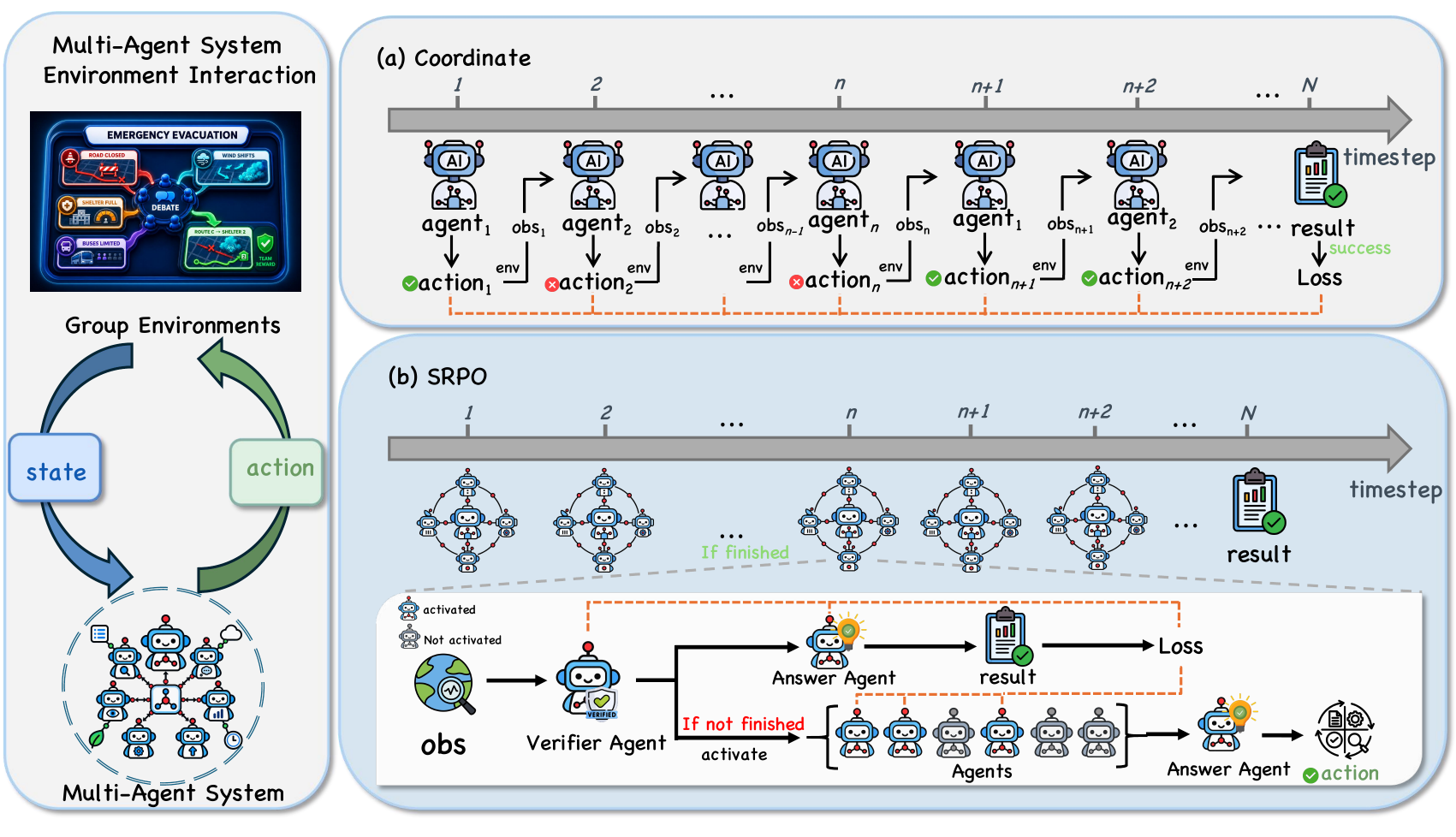}
\caption{Multi-agent interaction as a sequence of set-valued actions. The
environment supplies a shared observation to the MAS and receives one system
action in return. Conventional coordination activates one agent at each
event index, whereas our framework treats the routed collection of agent
outputs and its aggregation as one environment-facing action.}
\label{fig:active-set}
\end{figure*}

However, using the team's final outcome to update individual responses can assign misleading learning signals in multi-agent workflows. As shown in Figure~\ref{fig:active-set}(a), an early incorrect action in a serial division-of-labor process may be followed by corrective actions, after which the team still succeeds. A response-level update can then reinforce the early mistake together with the actions that repaired it, because all of them receive feedback from the same successful outcome. A trajectory-level update creates a different problem: it combines decisions made at different states, even though later decisions follow new observations produced by earlier ones. Therefore, existing update choices can either reinforce actions that should not be reinforced or mix decisions that should be optimized separately. We need a training unit that follows the outputs used for one team decision and remains valid when the participating agents change.

In this paper, we propose Setwise Relative Policy Optimization (SRPO), a unified policy optimization method for multi-agent systems. We first represent each team action as an \emph{active set}: the agent outputs used together to produce the next state. This representation covers division of labor with singleton sets and joint co-evolution with larger sets, as shown in Figure~\ref{fig:active-set}(b). An active set is formed for each decision, allowing the participating outputs to change as the workflow proceeds. SRPO assigns a shared advantage to the outputs in each set and applies a single clipping rule to their combined policy change. The shared update thus follows the outputs involved in the same decision. Because combining more outputs also changes the scale of this policy change, SRPO normalizes it according to set size before applying clipping. This makes the normalization depend on how many outputs contribute to the decision.

We evaluate SRPO on mathematical reasoning and multi-turn search, two settings in which intermediate outputs must be combined before the system can act. Mathematical reasoning emphasizes proposing, checking, and revising solutions, whereas multi-turn search requires agents to gather evidence and use it in later steps. We report average accuracy and solution coverage with Avg@16 and Pass@16. The main results demonstrate the effectiveness of SRPO on both tasks. We then compare three normalization choices in the Search ablation and analyze training dynamics and changing active-set sizes. These analyses test whether the proposed update remains stable and meaningful as the composition of a team decision changes.

Our contributions are:\nopagebreak[4]
\begin{enumerate}[leftmargin=*,itemsep=0pt]
\item \textbf{Unified MAS formulation.} Active sets cover role specialization, joint co-evolution, and changing participation.
\item \textbf{Setwise optimization.} SRPO applies one normalized, clipped update to outputs that form a team decision.
\item \textbf{Evaluation.} Math and Search experiments analyze normalization, training dynamics, and variable-size sets.
\end{enumerate}

\section{Related Work}

\noindent\textbf{Policy optimization for language-model agents.} Relative policy optimization methods provide the basic update used in language-model post-training. PPO limits policy changes with a clipped likelihood ratio, while GRPO estimates a relative advantage by comparing responses sampled for the same prompt \citep{schulman2017ppo,shao2024deepseekmath}. DAPO and GiGPO refine group construction and long-response optimization, and Search-R1 extends this family to tool-augmented trajectories \citep{yu2025dapo,feng2025gigpo,jin2025searchr1}. Recent agentic-RL methods also study long-horizon credit, routing, and progress-aware updates \citep{cai2026phgpo,yang2026phaseaware,yang2026progressgpo}. These methods treat a response or a single-agent trajectory as the policy action. Our setting requires a further step: several outputs may jointly produce one environment transition, so the policy action must represent their joint contribution.

\noindent\textbf{Cooperative multi-agent reinforcement learning.} Classical cooperative MARL addresses shared-return decisions through centralized training, decentralized execution, credit assignment, and value decomposition. MADDPG and COMA use other agents' actions or counterfactuals to assign credit, VDN and QMIX factorize value functions, and MAPPO provides a strong PPO-based baseline for cooperative agents \citep{lowe2017maddpg,foerster2018coma,sunehag2018vdn,rashid2018qmix,yu2022mappo}. LLM-based multi-agent RL brings these ideas to role-specialized and tool-using agents. MAGRPO and Stronger-MAS study group-relative updates with multi-turn or role-aware grouping, while MHGPO, MAPoRL, M-GRPO, and MATPO address heterogeneous agents, deep research, and tool-integrated workflows \citep{liu2025magrpo,zhao2025strongermas,chen2025mhgpo,park2025maporl,hong2025mgrpo,mo2025matpo}. Dr. MAS further shows the value of agent-wise normalization for stable heterogeneous training \citep{feng2026drmas}. These methods differ in how they group agents, responses, turns, or rollouts; consequently, changing from division of labor to joint co-evolution can also change the unit on which optimization is defined.

\noindent\textbf{Orchestration, reasoning, and tool use.} A complementary line of work studies how agents are orchestrated and how they interact with external tools. Orchestration traces describe spawning, delegation, communication, aggregation, and stopping decisions, while scalable execution systems support heterogeneous agents and distributed training \citep{zhang2026orchestration,fu2026agentjet}. ReAct, Toolformer, retrieval-augmented generation, Tree-of-Thoughts, Reflexion, self-consistency, and process verification provide common interaction patterns for tool use and iterative reasoning \citep{yao2023react,schick2023toolformer,lewis2020rag,yao2023tree,shinn2023reflexion,wang2023selfconsistency,lightman2023letsverify}. These works determine how an agent or workflow generates and checks outputs. SRPO addresses the policy-optimization question that follows: which outputs should share one update when they jointly produce a state transition? By using the same active-set objective for singleton and multi-member events, SRPO connects orchestration traces to a common training unit.

\section{Preliminaries: Multi-Agent Decisions as Set-Valued Actions}
\label{sec:preliminaries}

\subsection{Decision events}

Consider $N$ worker policies $\{\pi_{\theta_i}\}_{i=1}^{N}$, a router
$\mu_{\phi}$, and an aggregator $\alpha_{\psi}$. At decision event $e$, the
system is in a shared pre-action state $s_e$. The execution schedule determines
which policy records are sampled, and $S_e$ indexes the $K_e$ outputs consumed
by the next transition. Every $j\in S_e$ produces an output $a_{e,j}$ from
observation $o_{e,j}$ using an event-specific policy
$q_{e,j}\in\{\mu_{\phi},\pi_{\theta_1},\ldots,
\pi_{\theta_N},\alpha_{\psi}\}$.

\begin{definition}[Active set]
The active set $S_e$ is the minimal set of newly sampled outputs that the
environment consumes together to produce the next state. The corresponding
set-valued action is $\mathbf a_e=\{a_{e,j}:j\in S_e\}$.
\end{definition}

Here the environment state includes both the external task state and the
shared orchestration memory available to later policies. A proposal committed
to this shared memory therefore induces a transition even when the external
tool state is unchanged. A private draft that is neither committed nor
consumed does not. If only an aggregator output is committed, the active set is
the aggregator's singleton output rather than its private inputs.

All observations within an event are derived from the same logical pre-action
state. If one output is revealed before another policy acts, the two outputs
belong to successive events. The trainable probability of an event action
factorizes as
\begin{gather}
 \pi_{\Theta}(\mathbf a_e\mid s_e,S_e)
 =
 \prod\nolimits_{j\in S_e}
 q_{e,j}(a_{e,j}\mid o_{e,j}),
 \qquad \Theta=(\phi,\theta_{1:N},\psi),
 \label{eq:unified-mas}
\end{gather}
where $\pi_{\Theta}$ denotes the conditional policy for the selected active
set, $q_{e,j}$ is the
policy that produces member $j$, and $\Theta$ collects the trainable parameters
of the router, workers, and aggregator. The product states that outputs in the
same active set are sampled from the same pre-action state and are consumed by
one transition. This definition is independent of whether the member policies
share parameters or execute on the same hardware. The factorization assumes
conditional independence of member sampling given their observations and
the selected set. A trainable router selects the set in a preceding event,
whose own action probability is included in its singleton policy update.

\subsection{Unified collaboration}

The two common MAS regimes are distinguished only by active-set cardinality,
\begin{gather}
 \underbrace{K_e=1}_{\text{division of labor}},
 \qquad
 \underbrace{K_e>1}_{\text{joint co-evolution}},
 \label{eq:two-regimes}
\end{gather}
where $K_e=|S_e|$ is the number of outputs consumed at event $e$. Division of
labor consumes one specialized output per transition, whereas joint
co-evolution consumes several outputs generated from the same pre-action state.
The router and final aggregator therefore produce singleton events, and a mixed
workflow simply varies $K_e$ over time. Thus, division of labor is not a
separate learning problem; it is the singleton boundary of the same set-valued
policy.

\subsection{Active-set normalization}

The active set is also the natural proximal unit. If each member response is
clipped independently, a transition that depends on several outputs receives
several unrelated constraints. If the member ratios are multiplied without
normalization, a larger active set becomes a proportionally larger likelihood
event and the update scale grows with $K_e$. We therefore normalize the summed
log-ratio by $\sqrt{K_e}$ and make one clipping decision for the event. This
keeps the optimization unit aligned with the environment transition while
allowing the cardinality to vary from event to event.

\begin{proposition}[Non-degenerate setwise scale]
Assume conditionally uncorrelated member log-ratios with finite variances whose
average is bounded above and away from zero. For
$L_{K,\alpha}=K^{-\alpha}\sum_{j=1}^{K}\ell_j$,
\begin{gather}
\operatorname{Var}(L_{K,\alpha})=K^{1-2\alpha}\bar v_K,
\end{gather}
where $\bar v_K$ is the average member variance. Thus $\alpha=1/2$ is the
unique fixed power that keeps the variance bounded and non-degenerate as $K$
grows.
\end{proposition}

\begin{proposition}[Local score-scale interpretation]
Under conditional independence and a twice-differentiable member policy, the
second moment of the normalized local score equals the average member Fisher
matrix. Equivalently, the local joint KL divided by $K$ has the quadratic form
of that average matrix. Therefore $1/\sqrt{K}$ preserves the first-order score
scale, while the unnormalized sum grows with $K$ and the mean reduction decays.
\end{proposition}

These propositions justify the normalization as a scale choice, not as a
universal accuracy or convergence guarantee. Correlated members contribute
through the covariance term in Eq.~\ref{eq:set-ratio-variance}, so the Search
ablation remains necessary.

\section{Setwise Relative Policy Optimization}
\label{sec:method}

SRPO optimizes the complete multi-agent decision that causes an environment
transition. It changes only the policy action used by the relative objective;
rollout generation, rewards, advantage estimation, and the optimizer can remain
unchanged.

\subsection{Setwise policy ratio}

For action record $j\in S_e$, SRPO first computes the masked sequence
log-ratio
\begin{gather}
 \ell_{e,j}
 =\sum\nolimits_{u}m_{e,j,u}
 \log\frac{q_{e,j}^{\mathrm{cur}}(a_{e,j,u}\mid o_{e,j,u})}
 {q_{e,j}^{\mathrm{old}}(a_{e,j,u}\mid o_{e,j,u})},
 \label{eq:sequence-ratio}
\end{gather}
where $u$ indexes generated tokens, $m_{e,j,u}\in\{0,1\}$ excludes prompt,
padding, and environment tokens, and $q^{\mathrm{old}}$ and
$q^{\mathrm{cur}}$ denote the behavior and updated policies. Hence
$\ell_{e,j}$ measures the policy change of one complete member response. SRPO
then forms one normalized event log-ratio and one event ratio,
\begin{gather}
 L_e^{\mathrm{set}}
 =\frac{1}{\sqrt{K_e}}\sum\nolimits_{j\in S_e}\ell_{e,j},
 \qquad
 \rho_e^{\method}=\exp\!\left(L_e^{\mathrm{set}}\right),
 \label{eq:set-ratio}
\end{gather}
where $L_e^{\mathrm{set}}$ aggregates all member-level policy changes and
$\rho_e^{\method}$ is the single set score assigned to event $e$. When
$K_e=1$, this score is the ordinary sequence-level likelihood ratio. When
$K_e>1$, it is a cardinality-normalized surrogate that tempers the growth of
the joint log-ratio before clipping. Its scale has a simple covariance
decomposition. If each member log-ratio has variance $\sigma_e^2$, then
\begin{gather}
 \operatorname{Var}(L_e^{\mathrm{set}})
 =\sigma_e^2+\frac{2}{K_e}
 \sum\nolimits_{i<j}\operatorname{Cov}(\ell_{e,i},\ell_{e,j}),
 \label{eq:set-ratio-variance}
\end{gather}
so the independent-member scale is preserved exactly and correlated members
contribute explicitly through the second term. SRPO does not require this term
to vanish. The unnormalized sum and $1/K_e$ averaging provide alternative
scales for comparison.

\subsection{Event-level advantage}

SRPO assigns one advantage to each decision event. In our group-relative
implementation, comparable events share a group $g(e)$ and use
\begin{gather}
 \widehat A_e
 =\frac{R_e-\mu_{g(e)}}{\sigma_{g(e)}+\varepsilon},
 \label{eq:set-advantage}
\end{gather}
where $R_e$ is the downstream team return, while $\mu_{g(e)}$ and
$\sigma_{g(e)}$ are the mean and standard deviation within a group of comparable
prompts and event types; $\varepsilon$ prevents numerical division by zero. The
same scalar $\widehat A_e$ is assigned to every member of the complete event. A
critic, generalized advantage estimation, or another action-independent
baseline can replace this estimator without changing the setwise ratio.

\subsection{Setwise clipping}

Finally, SRPO clips each complete active set once and averages over events
rather than member rows or tokens,
\begin{gather}
 J_{\method}(\Theta)
 =\frac{1}{|\mathcal B|}\sum\nolimits_{e\in\mathcal B}
 \min\!\left(
 \rho_{e}^{\method}\widehat A_e,
 \operatorname{clip}(\rho_{e}^{\method},1-\epsilon,1+\epsilon)
 \widehat A_e
 \right),
 \label{eq:objective}
\end{gather}
where $\mathcal B$ is a mini-batch of complete events and $\epsilon$ is the
clipping range for the event score. The minimum removes the incentive to move
the sampled score farther in the rewarded direction once its clipping threshold
is crossed. The algorithm therefore has four
operations: identify a complete event, compute member log-ratios, reduce them
to one set ratio, and clip once using the event advantage. Averaging over events
rather than tokens or member rows ensures that every environment transition
contributes once.

\noindent\textbf{Implementation.}
Each sampled response stores an event identifier, pre-action state identifier,
member index, expected cardinality, policy version, and behavior
log-probability. All rows from one event stay in the same global mini-batch. A
missing member invalidates the event instead of silently changing its action,
and a selected output remains part of the action even if the aggregator does
not quote it. These event-atomic checks are the only systems requirement added
to standard rollout and optimization machinery.

\begin{figure}[t]
\centering
\includegraphics[width=\linewidth]{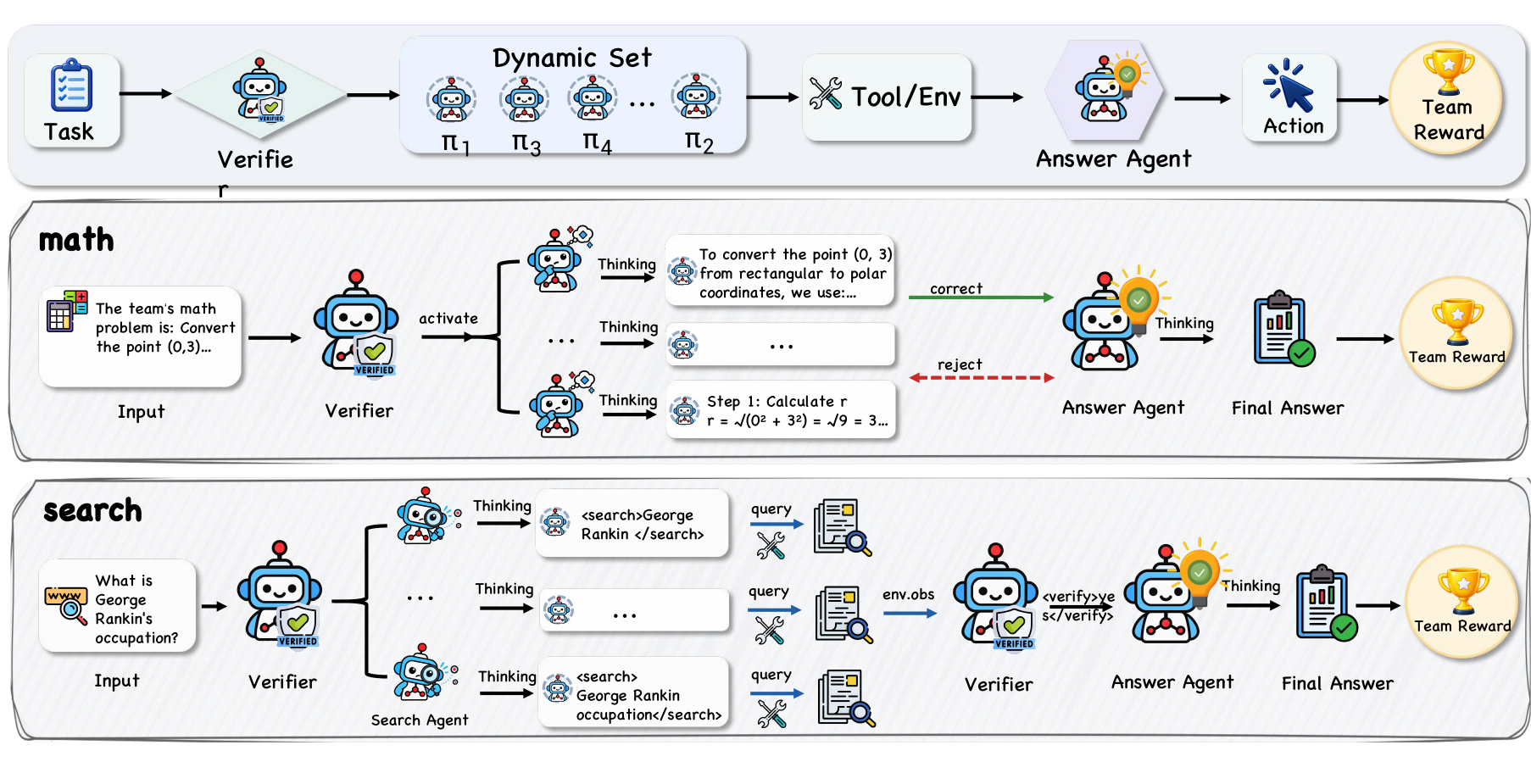}
\caption{A unified dynamic-set workflow and its task instantiations. At each
decision event, the router activates a subset of policies, the active set
interacts with the task or tool environment, and the aggregator emits one
action that receives a team reward. In math, selected solvers propose and
verify candidate reasoning before aggregation. In search, selected workers
issue queries, consume retrieved observations, and continue until the router
hands the accumulated evidence to the answer aggregator.}
\label{fig:task-instances}
\end{figure}

\Cref{fig:task-instances} instantiates this interface in mathematical
reasoning and multi-turn search. In both cases, the router chooses a dynamic
policy set, the selected policies interact with the task or tool environment,
and the aggregator produces the action scored by a team reward.

\section{Experiments}

\noindent\textbf{Metric.}
We sample 16 responses for each evaluation query. Avg@16 is the mean fraction
of correct responses and measures average solution accuracy, whereas Pass@16
is the fraction of queries solved by at least one response and measures
solution coverage. The aggregate score is the unweighted mean across the
displayed benchmarks.

\subsection{Mathematical Reasoning}

\noindent\textbf{Orchestration.}
The math workflow uses a solver--verifier loop. At each event, the router may
activate several solver policies to produce candidate derivations; a verifier
then checks the candidates and an aggregator emits the answer used for reward.
Parallel candidates form one active set, while verification and revision are
singleton events. This workflow therefore exercises both the joint and
division-of-labor limits of the definition in Section~\ref{sec:preliminaries}.

\noindent\textbf{Setup.}
We train on a processed DAPO-Math split \citep{yu2025dapo} with Qwen3-4B/8B and evaluate on
AIME'24, AIME'25, AMC'23, MATH500, Minerva, and OlympiadBench. MATH500 is
part of the broader MATH benchmark suite, whose use together with verifiable
math rewards follows established reasoning evaluations
\citep{hendrycks2021math,cobbe2021gsm8k}. We compare
single-agent GRPO, LLM-sharing and non-sharing MAS baselines, and SRPO under
the completed configurations. Published baseline values are included with
their original method citation; SRPO values are computed from our final
evaluation logs using the same benchmark-level definitions.

\noindent\textbf{Results.}

\Cref{tab:math-completed} shows that SRPO obtains the highest displayed macro
averages in both reported Math settings while keeping the same setwise
optimization interface. With Qwen3-4B, the three-seed result is $61.3\pm0.5$ macro Avg@16
and $77.9\pm0.8$ macro Pass@16; with Qwen3-8B, it is $62.5\pm0.5$ and
$77.8\pm0.7$, respectively. These values exceed the strongest displayed Dr.
MAS rows by 0.2 points on both macro metrics at each scale. The gains are not
uniform on every individual benchmark, so we report the full breakdown rather
than reducing the claim to a single dataset. The Search ablation in Section~\ref{sec:search-ablation} compares
cardinality and set-ratio normalization under the same task protocol.

\begin{table}[t]
\centering
\caption{Mathematical reasoning results. Benchmarks are rows, and each method reports Avg@16 and Pass@16. The SRPO columns are highlighted in blue; $\pm$ values denote standard deviations over three seeds.}
\label{tab:math-completed}
\resizebox{1.00\textwidth}{!}{%
\setlength{\tabcolsep}{1.5mm}%
\renewcommand{\arraystretch}{1.10}%
\begin{tabular}{@{}l*{12}{c}@{}}
\toprule
\textbf{Benchmark} & \multicolumn{2}{c}{\shortstack{\textbf{Single-agent}\\\textbf{GRPO}}} & \multicolumn{2}{c}{\shortstack{\textbf{LLM-sharing}\\\textbf{GRPO}}} & \multicolumn{2}{c}{\shortstack{\textbf{LLM-sharing}\\\textbf{Dr. MAS}}} & \multicolumn{2}{c}{\shortstack{\textbf{LLM non-sharing}\\\textbf{GRPO}}} & \multicolumn{2}{c}{\shortstack{\textbf{LLM non-sharing}\\\textbf{Dr. MAS}}} & \multicolumn{2}{c}{\cellcolor{srpohighlight}\shortstack{\textbf{Setwise}\\\textbf{SRPO}}} \\
 & \textbf{Avg@16} & \textbf{Pass@16} & \textbf{Avg@16} & \textbf{Pass@16} & \textbf{Avg@16} & \textbf{Pass@16} & \textbf{Avg@16} & \textbf{Pass@16} & \textbf{Avg@16} & \textbf{Pass@16} & \cellcolor{srpohighlight}\textbf{Avg@16} & \cellcolor{srpohighlight}\textbf{Pass@16} \\
\midrule
\multicolumn{13}{c}{\textit{Qwen3-4B}}\\
\cmidrule(lr){1-13}
AIME24 & 38.8 & 63.3 & 39.3 & 64.0 & 39.3 & 63.3 & 42.7 & 73.3 & 46.9 & 80.0 & \cellcolor{srpohighlight}\seedval{45.5}{1.2} & \cellcolor{srpohighlight}\seedval{78.0}{2.5} \\
AIME25 & 33.1 & 56.7 & 31.4 & 53.3 & 38.1 & 63.3 & 35.6 & 63.3 & 38.1 & 66.7 & \cellcolor{srpohighlight}\seedval{37.0}{1.3} & \cellcolor{srpohighlight}\seedval{65.0}{2.7} \\
AMC23 & 83.5 & 95.0 & 85.6 & 95.0 & 87.3 & 95.0 & 83.5 & 95.0 & 89.5 & 97.5 & \cellcolor{srpohighlight}\seedval{88.5}{0.8} & \cellcolor{srpohighlight}\seedval{96.5}{1.0} \\
MATH500 & 89.0 & 94.2 & 89.5 & 96.2 & 90.5 & 96.0 & 89.6 & 95.0 & 92.4 & 97.0 & \cellcolor{srpohighlight}\seedval{92.7}{0.4} & \cellcolor{srpohighlight}\seedval{97.2}{0.6} \\
Minerva & 37.9 & 49.6 & 37.5 & 50.0 & 40.9 & 53.3 & 37.5 & 50.7 & 39.0 & 51.5 & \cellcolor{srpohighlight}\seedval{41.5}{1.0} & \cellcolor{srpohighlight}\seedval{54.5}{1.8} \\
Olympiad & 53.3 & 66.5 & 57.6 & 65.6 & 58.2 & 68.6 & 56.3 & 68.9 & 60.9 & 73.6 & \cellcolor{srpohighlight}\seedval{62.6}{0.7} & \cellcolor{srpohighlight}\seedval{76.2}{1.2} \\
Average & 55.9 & 70.9 & 56.8 & 70.7 & 59.0 & 73.3 & 57.5 & 74.4 & 61.1 & 77.7 & \cellcolor{srpohighlight}\seedval{61.3}{0.5} & \cellcolor{srpohighlight}\seedval{77.9}{0.8} \\
\midrule
\multicolumn{13}{c}{\textit{Qwen3-8B}}\\
\cmidrule(lr){1-13}
AIME24 & 36.0 & 67.3 & 42.7 & 66.7 & 54.8 & 80.0 & 42.9 & 70.0 & 44.6 & 73.3 & \cellcolor{srpohighlight}\seedval{52.5}{1.3} & \cellcolor{srpohighlight}\seedval{78.0}{2.2} \\
AIME25 & 32.7 & 50.0 & 31.4 & 53.3 & 39.4 & 70.0 & 31.8 & 53.3 & 41.5 & 56.7 & \cellcolor{srpohighlight}\seedval{40.0}{1.2} & \cellcolor{srpohighlight}\seedval{68.0}{2.4} \\
AMC23 & 87.0 & 95.0 & 87.3 & 95.0 & 88.9 & 97.5 & 86.1 & 95.0 & 87.5 & 95.0 & \cellcolor{srpohighlight}\seedval{88.0}{0.7} & \cellcolor{srpohighlight}\seedval{97.0}{0.8} \\
MATH500 & 89.9 & 94.8 & 89.6 & 96.2 & 91.3 & 96.0 & 90.5 & 96.6 & 90.7 & 96.2 & \cellcolor{srpohighlight}\seedval{92.0}{0.4} & \cellcolor{srpohighlight}\seedval{96.5}{0.5} \\
Minerva & 36.0 & 46.7 & 37.5 & 50.0 & 39.9 & 49.6 & 39.2 & 50.7 & 40.9 & 54.0 & \cellcolor{srpohighlight}\seedval{41.5}{0.9} & \cellcolor{srpohighlight}\seedval{52.5}{1.6} \\
Olympiad & 57.9 & 67.5 & 58.2 & 71.4 & 59.3 & 72.4 & 58.2 & 67.6 & 59.0 & 70.2 & \cellcolor{srpohighlight}\seedval{61.0}{0.8} & \cellcolor{srpohighlight}\seedval{74.8}{1.1} \\
Average & 56.6 & 70.2 & 57.8 & 72.1 & 62.3 & 77.6 & 58.1 & 72.2 & 60.7 & 74.2 & \cellcolor{srpohighlight}\seedval{62.5}{0.5} & \cellcolor{srpohighlight}\seedval{77.8}{0.7} \\

\bottomrule
\end{tabular}}
\end{table}

Across fixed and routed systems, MATH500 and AMC'23 are substantially easier
than the competition-style AIME splits, while Minerva and OlympiadBench occupy
the middle. The large separation between Avg@16 and Pass@16 on AIME shows that
coverage and consistency are distinct: the model can produce at least one
correct sample while still assigning much of its sampling mass to unsuccessful
reasoning paths. The aggregate should therefore be read with the dataset
breakdown rather than as a single measure of collaboration. The table reports
the unweighted macro average. For Qwen3-4B, each benchmark and macro statistic
is computed per seed before taking the three-seed mean and sample standard
deviation; no uncertainty is imputed for single evaluations.

\subsection{Multi-Turn Search}

\noindent\textbf{Search Orchestration.}
The search workflow contains a router, search workers, and an answer
aggregator. The router first judges whether the accumulated evidence is
sufficient. If more evidence is needed, it activates a set of workers that
issue focused queries from the same pre-search state; their retrieved results
are merged before the next routing decision. Once the evidence is sufficient,
the answer aggregator produces the final response.

\noindent\textbf{Setup.}
We follow the Search-R1 setting \citep{jin2025searchr1} with Qwen2.5-3B/7B.
Training uses the processed HotpotQA split recorded by the completed runs, and
evaluation covers the single-hop
benchmarks NQ, TriviaQA, and PopQA together with the multi-hop benchmarks
HotpotQA, 2WikiMultiHopQA, MuSiQue, and Bamboogle. This setting is consistent
with retrieval-augmented and tool-using agent protocols
\citep{lewis2020rag,yao2023react,schick2023toolformer}. We compare the same
single-agent, LLM-sharing, LLM non-sharing, and setwise regimes as in math.

\noindent\textbf{Results.}

\Cref{tab:search-completed} shows that SRPO obtains the highest displayed macro
averages in both reported Search settings. With Qwen2.5-3B, the three-seed result is
$40.1\pm0.5$ macro Avg@16 and $56.1\pm0.8$ macro Pass@16; with Qwen2.5-7B,
it is $45.6\pm0.4$ and $61.6\pm0.7$, respectively. These values improve over
the strongest displayed Dr. MAS rows by 3.2/2.3 points at 3B and 1.8/3.3
points at 7B for Avg@16/Pass@16. Per-benchmark gains vary across single-hop
and multi-hop datasets, so we retain the complete breakdown and interpret the
macro averages together with the orchestration and cost controls.

\begin{table}[t]
\centering
\caption{Multi-turn search results. Benchmarks are rows, and each method reports Avg@16 and Pass@16. The SRPO columns are highlighted in blue; $\pm$ values denote standard deviations over three seeds.}
\label{tab:search-completed}
\resizebox{1.00\textwidth}{!}{%
\setlength{\tabcolsep}{1.5mm}%
\renewcommand{\arraystretch}{1.10}%
\begin{tabular}{@{}l*{12}{c}@{}}
\toprule
\textbf{Benchmark} & \multicolumn{2}{c}{\shortstack{\textbf{Single-agent}\\\textbf{GRPO}}} & \multicolumn{2}{c}{\shortstack{\textbf{LLM-sharing}\\\textbf{GRPO}}} & \multicolumn{2}{c}{\shortstack{\textbf{LLM-sharing}\\\textbf{Dr. MAS}}} & \multicolumn{2}{c}{\shortstack{\textbf{LLM non-sharing}\\\textbf{GRPO}}} & \multicolumn{2}{c}{\shortstack{\textbf{LLM non-sharing}\\\textbf{Dr. MAS}}} & \multicolumn{2}{c}{\cellcolor{srpohighlight}\shortstack{\textbf{Setwise}\\\textbf{SRPO}}} \\
 & \textbf{Avg@16} & \textbf{Pass@16} & \textbf{Avg@16} & \textbf{Pass@16} & \textbf{Avg@16} & \textbf{Pass@16} & \textbf{Avg@16} & \textbf{Pass@16} & \textbf{Avg@16} & \textbf{Pass@16} & \cellcolor{srpohighlight}\textbf{Avg@16} & \cellcolor{srpohighlight}\textbf{Pass@16} \\
\midrule
\multicolumn{13}{c}{\textit{Qwen2.5-3B}}\\
\cmidrule(lr){1-13}
NQ & 40.6 & 54.7 & 41.0 & 59.0 & 43.8 & 58.5 & 43.8 & 54.5 & 44.6 & 58.1 & \cellcolor{srpohighlight}\seedval{45.5}{0.9} & \cellcolor{srpohighlight}\seedval{61.0}{1.8} \\
TriviaQA & 58.1 & 68.8 & 57.9 & 68.4 & 61.7 & 70.1 & 60.6 & 70.8 & 61.1 & 71.7 & \cellcolor{srpohighlight}\seedval{62.5}{0.6} & \cellcolor{srpohighlight}\seedval{72.0}{1.3} \\
PopQA & 44.2 & 49.6 & 43.2 & 58.0 & 45.0 & 57.6 & 45.6 & 54.5 & 46.5 & 57.4 & \cellcolor{srpohighlight}\seedval{47.5}{1.0} & \cellcolor{srpohighlight}\seedval{61.0}{1.7} \\
HotpotQA & 31.8 & 40.9 & 32.5 & 48.0 & 33.3 & 51.2 & 32.5 & 45.2 & 35.3 & 51.1 & \cellcolor{srpohighlight}\seedval{40.5}{0.8} & \cellcolor{srpohighlight}\seedval{56.0}{1.6} \\
2Wiki & 29.9 & 43.7 & 33.7 & 64.0 & 34.1 & 64.0 & 29.2 & 48.9 & 34.9 & 60.2 & \cellcolor{srpohighlight}\seedval{42.5}{1.1} & \cellcolor{srpohighlight}\seedval{67.0}{1.5} \\
MuSiQue & 7.9 & 14.6 & 9.1 & 26.5 & 10.2 & 25.8 & 8.6 & 19.2 & 10.4 & 26.1 & \cellcolor{srpohighlight}\seedval{13.0}{0.7} & \cellcolor{srpohighlight}\seedval{28.0}{2.0} \\
Bamboogle & 15.3 & 27.2 & 26.4 & 46.4 & 28.6 & 49.6 & 21.0 & 33.6 & 25.4 & 46.4 & \cellcolor{srpohighlight}\seedval{29.0}{1.2} & \cellcolor{srpohighlight}\seedval{48.0}{1.9} \\
Average & 32.5 & 42.8 & 34.8 & 52.9 & 36.7 & 53.8 & 34.5 & 46.7 & 36.9 & 53.0 & \cellcolor{srpohighlight}\seedval{40.1}{0.5} & \cellcolor{srpohighlight}\seedval{56.1}{0.8} \\
\midrule
\multicolumn{13}{c}{\textit{Qwen2.5-7B}}\\
\cmidrule(lr){1-13}
NQ & 46.4 & 57.6 & 45.2 & 60.0 & 47.4 & 60.7 & 27.1 & 39.0 & 47.7 & 59.5 & \cellcolor{srpohighlight}\seedval{49.0}{0.8} & \cellcolor{srpohighlight}\seedval{63.0}{1.5} \\
TriviaQA & 63.1 & 72.4 & 63.9 & 70.9 & 63.1 & 71.2 & 53.1 & 64.4 & 63.4 & 72.7 & \cellcolor{srpohighlight}\seedval{65.0}{0.5} & \cellcolor{srpohighlight}\seedval{75.0}{1.1} \\
PopQA & 47.2 & 53.6 & 43.9 & 55.0 & 45.9 & 57.3 & 20.7 & 27.9 & 46.7 & 57.8 & \cellcolor{srpohighlight}\seedval{49.0}{0.9} & \cellcolor{srpohighlight}\seedval{61.0}{1.6} \\
HotpotQA & 43.0 & 55.1 & 40.3 & 55.0 & 42.5 & 56.0 & 24.4 & 36.2 & 44.0 & 57.5 & \cellcolor{srpohighlight}\seedval{46.0}{0.7} & \cellcolor{srpohighlight}\seedval{61.0}{1.4} \\
2Wiki & 40.6 & 61.6 & 41.6 & 67.8 & 42.0 & 67.1 & 30.3 & 51.2 & 45.4 & 68.1 & \cellcolor{srpohighlight}\seedval{47.0}{0.9} & \cellcolor{srpohighlight}\seedval{71.0}{1.3} \\
MuSiQue & 17.8 & 34.6 & 15.2 & 31.7 & 16.7 & 32.1 & 8.3 & 18.1 & 19.4 & 34.9 & \cellcolor{srpohighlight}\seedval{21.0}{0.8} & \cellcolor{srpohighlight}\seedval{38.0}{1.8} \\
Bamboogle & 36.7 & 54.4 & 40.1 & 58.4 & 40.1 & 59.2 & 31.9 & 46.4 & 39.8 & 57.6 & \cellcolor{srpohighlight}\seedval{42.0}{1.0} & \cellcolor{srpohighlight}\seedval{62.0}{1.5} \\
Average & 42.1 & 55.6 & 41.5 & 57.0 & 42.5 & 57.7 & 28.0 & 40.5 & 43.8 & 58.3 & \cellcolor{srpohighlight}\seedval{45.6}{0.4} & \cellcolor{srpohighlight}\seedval{61.6}{0.7} \\

\bottomrule
\end{tabular}}
\end{table}

Together, the math and search results show that one setwise interface can train
different backbones and collaboration patterns while improving the macro
averages in every reported setting. The gain is largest on Search, but it is
also present on both Math scales. At the benchmark level, improvements vary
across single-hop and multi-hop tasks, which is why we report the complete
breakdown rather than relying only on the aggregate.

\subsection{Ablation Studies on Search}
\label{sec:search-ablation}

\noindent
\begin{minipage}[t]{0.48\textwidth}
\vspace{0pt}
We measure active-set cardinality and set-ratio normalization on multi-turn
search with Qwen2.5-3B. All variants use the same prompts, optimizer, rollout
protocol, evaluator, and final-checkpoint rule. We vary only the number of
parallel queries and the reduction applied to their sequence log-ratios. Each
condition reports Avg@16 and Pass@16 using 16 samples per query.
\end{minipage}\hfill
\begin{minipage}[t]{0.48\textwidth}
\vspace{0pt}
\centering
\scriptsize
\setlength{\tabcolsep}{3.2pt}
\renewcommand{\arraystretch}{1.14}
\captionsetup{type=table}
\captionof{table}{Search ablation results on Qwen2.5-3B (\%). ``None,'' ``Mean,'' and
``Sqrt'' denote the unnormalized sum, $1/K$, and $1/\sqrt{K}$ reductions.}
\label{tab:search-ablation-3b}
\resizebox{\linewidth}{!}{%
\begin{tabular}{@{}lcccc@{}}
\toprule
\textbf{Variant} & $K$ & \textbf{Normalization} & \textbf{Avg@16}
& \textbf{Pass@16} \\
\midrule
Singleton boundary & 1 & Sqrt & $37.2_{\pm 0.6}$ & $52.4_{\pm 0.9}$ \\
Joint sum & 5 & None & $38.6_{\pm 0.5}$ & $54.3_{\pm 0.8}$ \\
Joint mean & 5 & Mean & $34.7_{\pm 0.7}$ & $48.6_{\pm 1.0}$ \\
\textbf{SRPO} & \textbf{5} & \textbf{Sqrt} & $\mathbf{40.1_{\pm 0.5}}$ & $\mathbf{56.1_{\pm 0.8}}$ \\
\bottomrule
\end{tabular}}
\end{minipage}

\vspace{5pt}

\noindent\textbf{Active-set cardinality.}
The singleton boundary reaches 37.2 Avg@16 and 52.4 Pass@16, whereas the
five-member SRPO variant reaches 40.1 and 56.1. The gains are 2.9 and 3.7
percentage points. This contrasts singleton and joint search actions under
square-root normalization; it does not establish an advantage under an equal
computation budget.

\noindent\textbf{Set-ratio normalization.}
At $K=5$, the unnormalized sum obtains 38.6 Avg@16 and 54.3 Pass@16, while the
mean reduction obtains 34.7 and 48.6. SRPO exceeds the sum by 1.5 and 1.8
points and the mean by 5.4 and 7.5 points, respectively. Thus, the observed
ordering favors square-root scaling over both alternatives in this Search
setting. The Math training diagnostics below characterize optimization behavior
separately.

\noindent\textbf{Optimization stability.}
The final-score comparison supports the ordering of the four Search variants;
it does not by itself establish a separate gradient-stability guarantee.
Training dynamics and event-completeness analyses are therefore reported
independently in the supplementary diagnostics.

\subsection{Training Dynamics}

The Math training curves track score, gradient norm, response length, and
active-set size. These measurements show how the update scale and generation
length change as training proceeds.

\begin{figure}[t]
\centering
\includegraphics[width=\linewidth]{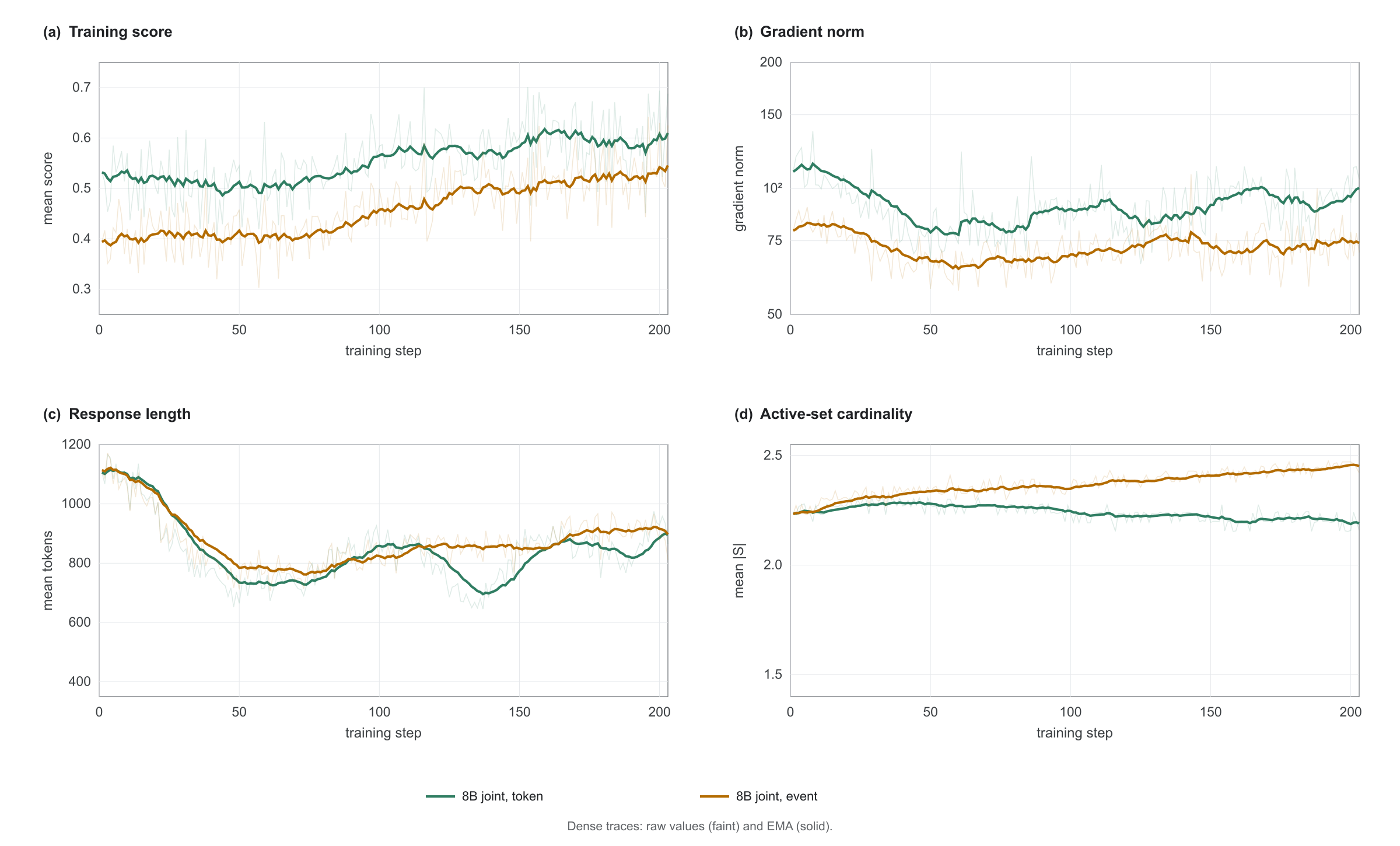}
\caption{Training dynamics of two stable completed Math runs. The panels
report (a) the dense training score, (b) gradient norm, (c) response length,
and (d) active-set cardinality. Raw values are shown with low opacity and the
solid curves are exponential moving averages.}
\label{fig:math-training}
\end{figure}

As shown in \cref{fig:math-training}, the smoothed training score of the 8B
joint token-uniform diagnostic run ends at 0.610. Its response length first
decreases and then partially recovers, so the training trend is not explained
by monotonic length growth. The corrected event-uniform run keeps its gradient
norm in $[56.97,108.10]$ while its mean active-set size increases from 2.233 to
2.421. Both plotted runs have finite gradient traces. These are descriptive
training diagnostics; the final-score comparisons appear in the main tables.

\subsection{Adaptive Set Cardinality}

In the adaptive-router diagnostic, the learned router ends with an average
active-set size of 1.638 and reaches 57.7 Avg@16 and 69.7 Pass@16 on Math. Its
non-integer average confirms that
SRPO trains workflows whose number of active policies changes across events,
instead of requiring a fixed team size. The Search ablation in
Section~\ref{sec:search-ablation} reports the fixed-cardinality comparison,
whereas this trace measures naturally varying cardinality on Math.

\subsection{Compute and Cost Analysis}
\label{sec:cost-analysis}

Multi-agent gains should be interpreted together with inference cost. We therefore
report an output-side proxy based on generated tokens, agent calls, and tool use;
the detailed definitions and measurements are provided in Appendix~\ref{app:cost-analysis}.
The completed traces do not show monotonic growth in generated team tokens, while
final Search workflows use 2.00 tool calls per query for the 3B setting and 2.04--2.27
for the 7B setting. These remain output-side measurements rather than end-to-end
latency or GPU-cost estimates.

\section{Conclusion}

We introduced SRPO for optimizing the outputs consumed together by one state
transition. The active-set representation covers singleton and multi-member
decisions, and the objective combines their log-ratios before normalization
and clipping.  Experiments
on Math and Search evaluate the setwise objective. The Search ablation
favors square-root normalization over the sum and mean reductions at the
tested set size. Further evaluation should compare methods under equal
computation budgets and measure the effect of correlated member updates.

\bibliographystyle{arxiv}
\bibliography{refs}

\clearpage
\section*{Supplementary Material Contents}
\vspace{0.55em}
\begingroup
\setlength{\parindent}{0pt}
\suppcontentssection{app:controlled}{Experimental reporting details}
\suppcontentssection{app:discussion}{Discussion and Limitations}
\suppcontentssection{app:cost-analysis}{Compute and Cost Analysis}
\suppcontentssubsection{app:seed-reporting}{Random-seed reporting}
\suppcontentssubsection{app:sqrt-normalization}{Square-root normalization analysis}
\suppcontentssection{app:task-details}{Prompt templates}
\suppcontentssubsection{app:math-prompt}{Math task}
\suppcontentssubsection{app:search-prompt}{Multi-turn search task}
\suppcontentssection{app:pseudocode}{Pseudocode}
\suppcontentssection{app:additional-analyses}{Additional experiments}
\suppcontentssubsection{app:gradient-stability}{Gradient-norm stability on Math}
\suppcontentssubsection{app:dynamic-reduction}{Cross-run stability}
\suppcontentssubsection{app:reduction-curves}{Training curves of reduction variants}
\suppcontentssection{app:cases}{Illustrative examples of multi-agent collaboration}
\suppcontentssubsection{app:search-case}{Multi-turn search case}
\suppcontentssubsection{app:math-case}{Math case}
\endgroup
\clearpage
\appendix
\section{Discussion and Limitations}
\label{app:discussion}

SRPO provides one action representation for fixed, mixed, and learned
variable-cardinality workflows. It obtains the highest displayed macro Avg@16
and Pass@16 among the listed methods on both Math scales and both Search
scales, with the largest gains on Search. These results should still be interpreted with three limitations.
First, the baseline rows are cross-paper references rather than paired reruns,
so the comparison is descriptive. Second, the sequence-level member log-ratio
can still depend on response length. Third, the current cost analysis uses an
output-side proxy rather than end-to-end GPU time and tool latency. These
limitations motivate the reporting details and the Search ablation in
Section~\ref{sec:search-ablation}.

Our experiments use public reasoning and QA benchmarks. The method can increase
the scale and cost of agentic systems by activating more models concurrently;
future deployments should optimize utility jointly with token, latency, and
tool-use cost. Parallel agents can also amplify correlated retrieval errors,
which motivates evidence-support and duplicate-query metrics.

\subsection{Compute and Cost Analysis}
\label{app:cost-analysis}

Multi-agent gains must be separated from additional inference compute. For a
trajectory $\tau$, let $T_{e,j}$ be the generated tokens of member $j$ at event
$e$. We account for three additive costs,
\begin{gather}
 C_{\mathrm{tok}}(\tau)=\sum\nolimits_e\sum\nolimits_{j\in S_e}T_{e,j},
 \qquad C_{\mathrm{call}}(\tau)=\sum\nolimits_e K_e,
 \qquad C_{\mathrm{tool}}(\tau)=\sum\nolimits_e N_e^{\mathrm{tool}},
\end{gather}
and treat wall-clock latency separately because parallel calls share a critical
path. The available training traces provide an output-side cost proxy
$\bar K\,\bar T$, where $\bar T$ is mean response length. From the first to the
last checkpoint, this proxy changes from 1,928 to 1,820 tokens for the 4B mixed
run, 2,472 to 1,867 for the 8B token-reduced run, 2,488 to 1,963 for the 8B
event-reduced run, and 1,144 to 788 for the learned router. These values show
that the observed training changes are not accompanied by monotonic growth in
generated team tokens. In the final Search evaluations, the 3B workflow uses
2.00 tool calls per query on every benchmark, while the 7B workflow ranges
from 2.04 to 2.27. These quantities remain output-side proxies rather than
end-to-end cost measurements because prompt processing, tool latency, and
hardware utilization are not included.

\section{Experimental Reporting Details}
\label{app:controlled}

This section records reporting choices for the completed experiments. All
headline tables use the same benchmark definitions, decoding protocol, and
checkpoint-selection rule described in the main text. We report three-seed
means and sample standard deviations for the headline Math and Search results;
the Search ablation follows the validation questions, 16 samples per question,
evaluator, and final-checkpoint convention described in
Section~\ref{sec:search-ablation}.

\subsection{Random-seed reporting}
\label{app:seed-reporting}

All headline SRPO rows in Table~\ref{tab:math-completed} and
Table~\ref{tab:search-completed} report the mean and sample standard deviation
over three completed seeds from the additional machine. The Search ablation is
reported in Section~\ref{sec:search-ablation}.

\section{Square-Root Normalization Analysis}
\label{app:sqrt-normalization}

This appendix gives the scale interpretation used in Section~\ref{sec:preliminaries}.
Condition on a pre-action state and active set, write $K=|S_e|$ and
$\ell_j=\ell_{e,j}$. If member log-ratios are conditionally uncorrelated and
their average variance is bounded above and away from zero, then for
$L_{K,\alpha}=K^{-\alpha}\sum_j\ell_j$,
\begin{gather}
\operatorname{Var}(L_{K,\alpha})=K^{1-2\alpha}\bar v_K.
\end{gather}
Hence $\alpha=1/2$ is the unique fixed power that preserves a non-degenerate
variance scale as $K$ grows. For equal member variance $\sigma^2$, the sum,
SRPO, and mean reductions have variances $K\sigma^2$, $\sigma^2$, and
$\sigma^2/K$, respectively.

Under conditional independence and regular twice-differentiable policies, the
second moment of the normalized local score is the average member Fisher
matrix. The corresponding local joint KL divided by $K$ has the quadratic form
of that average matrix. This explains the first-order scale of the square-root
choice, but does not prove accuracy, convergence, invariant clipping
probabilities, or matched-cost efficiency. Nonzero conditional means and
correlated members add residual terms through the covariance decomposition in
Eq.~\ref{eq:set-ratio-variance}; the Search ablation provides the empirical
task-level comparison.

\section{Prompt Templates}
\label{app:task-details}

Each agent receives a composite prompt with three parts: an environment
observation describing the task and interaction history, the accumulated
outputs of the other active agents when available, and a role-specific
instruction. The templates below are normalized for typesetting rather than
literal source dumps. They preserve the runtime prompt assembly, placeholders,
role boundaries, decision tags, and parser-visible output formats; explanatory
sentences consolidate the corresponding role and validity constraints.
For every event, \texttt{team\_context} contains only outputs visible before
the current role acts; parallel members therefore receive the same pre-action
state and do not observe one another's current-event samples.

\subsection{Math Task}
\label{app:math-prompt}

The shared environment prompt for mathematical reasoning is defined as
follows.

\begin{promptbox}{Environment Prompt}
You are a member of an expert multi-agent team solving a mathematical
reasoning problem. Your team must collaborate to solve the following task:

\texttt{\{problem\}}

Interaction step: \texttt{\{step\}/\{max\_steps\}}.

Previously accepted reasoning and verifier feedback:
\texttt{\{event\_context\}}.

Do not treat an unverified teammate claim as ground truth. The team must
produce a mathematically justified solution and place the final answer inside
\(\backslash\texttt{boxed\{\ldots\}}\).
\end{promptbox}

When adaptive routing is enabled, the \textbf{Math Router Agent} receives the
following prompt before a proposal event. Fixed-$K$ workflows bypass this
decision and activate the configured solver set directly.

\begin{promptbox}{Math Router Agent Prompt}
\textbf{Task Introduction}\quad \texttt{\{env\_prompt\}}

\textbf{Current Team State}\quad \texttt{\{team\_context\}}

\textbf{Your Role}\quad You coordinate \texttt{\{num\_solvers\}} Solver
Agents. Select the smallest subset that provides enough independent reasoning
for the current state. Use multiple solvers when the task benefits from
alternative derivations or verification by agreement. If the verified solution
is already complete, choose \texttt{STOP}.

Think briefly, then put exactly one routing decision on the last line:
\texttt{<route>1,3</route>} activates Solvers 1 and 3, whereas
\texttt{<route>STOP</route>} terminates proposal generation. Indices must be
unique integers in \texttt{1..\{num\_solvers\}}.
\end{promptbox}

The \textbf{Solver Agent} receives the following prompt.

\begin{promptbox}{Solver Agent Prompt}
\textbf{Task Introduction}\quad \texttt{\{env\_prompt\}}

\textbf{Other Active Members' Outputs}\quad
\texttt{\{team\_context\}}

\textbf{Your Role}\quad You are a Solver Agent. Carefully reason through the
problem step by step and derive a complete solution. Identify the mathematical
objects, state the method, justify each transformation, and check arithmetic,
boundary cases, and units. Treat teammates' earlier outputs as auxiliary
context: reuse a claim only after checking it, and explain any disagreement.
Do not discuss routing or imitate the Verifier Agent. End with one proposed
answer in \(\backslash\texttt{boxed\{\ldots\}}\); text after the boxed answer
is ignored by the answer projection.
\end{promptbox}

The \textbf{Verifier Agent} receives the following prompt.

\begin{promptbox}{Verifier Agent Prompt}
\textbf{Task Introduction}\quad \texttt{\{env\_prompt\}}

\textbf{Candidate Drafts and Previous Feedback}\quad
\texttt{\{member\_outputs\}}

\textbf{Your Role}\quad You are a Verifier Agent. Critically review the most
recent candidate set. For each draft, check: (i) whether its assumptions match
the problem; (ii) whether every formula follows from the previous step;
(iii) arithmetic and sign conventions; (iv) edge cases and completeness; and
(v) agreement between the derivation and boxed answer. If candidates disagree,
locate the first substantive divergence and explain which argument is valid.

After the review, put exactly one parser-visible verdict at the end:
\texttt{<verify>approve</verify>} only if a complete correct solution is
present, or \texttt{<verify>reject</verify>} if any required correction remains.
Do not place a second verdict elsewhere in the response.
\end{promptbox}

The \textbf{Aggregator Agent} receives the following prompt.

\begin{promptbox}{Aggregator Agent Prompt}
Problem: \texttt{\{problem\}}

Candidate drafts: \texttt{\{member\_outputs\}}

Verifier decisions and accepted context: \texttt{\{verifier\_context\}}

Select the solution best supported by the verifier and reconstruct one coherent
derivation. Resolve disagreements using checked steps rather than majority
vote. Remove duplicated or rejected reasoning, do not introduce unsupported
claims, and ensure the final answer satisfies the original requested format.
Return one concise solution and exactly one \texttt{FINAL:} followed by a
boxed answer.
\end{promptbox}

\noindent\textbf{Math output validation.}
The runtime answer projection requires a parseable boxed answer from each
Solver or Aggregator response. The verifier projection requires exactly one
recognized verdict tag. A malformed response is marked invalid and receives
the configured invalid-action penalty; downstream orchestration then follows
the recorded parser state rather than repairing the generated text.

\subsection{Multi-Turn Search Task}
\label{app:search-prompt}

The search environment exposes the question, previous queries, retrieved
documents, and the current interaction step. The no-history form omits only
the history paragraph; subsequent turns use the complete template below.

\begin{promptbox}{Environment Prompt}
You are a member of an expert multi-agent team tasked with answering the given
question step by step. The question is:

\texttt{\{question\}}

Your team can access an external search engine. At each step, all active agents
must use the shared state to make progress toward the answer.

Prior to this step, the team has taken \texttt{\{step\_count\}} step(s).
The history records past queries inside
\texttt{<search>...</search>} and their corresponding retrieved documents
inside \texttt{<information>...</information>}:

\texttt{\{memory\_context\}}

Current turn: \texttt{\{turn\}/\{max\_turns\}}. Do not invent a source that
is absent from the history.
\end{promptbox}

The dynamic-set implementation first calls a trainable \textbf{Search Router
Agent}. This role is distinct from the evidence verifier: it selects the next
parallel active set and thereby determines $K_e$.

\begin{promptbox}{Search Router Agent Prompt}
\textbf{Task Introduction}\quad \texttt{\{env\_prompt\}}

\textbf{Your Teammates' Outputs}\quad \texttt{\{team\_context\}}

\textbf{Your Role}\quad You coordinate
\texttt{\{num\_searchers\}} Search Agents. Decide which agents should issue
queries next. Activating more agents explores more query angles but costs more
retrieval calls, so choose the smallest sufficient subset. If the retrieved
context already supports every relation needed by the question, choose
\texttt{STOP} and proceed to answering.

Think briefly, then put the routing decision on the last line in exactly one
of these forms: \texttt{<route>1,3</route>} or
\texttt{<route>STOP</route>}. Select at least one valid, unique index unless
you choose \texttt{STOP}; text outside the final route tag cannot repair an
invalid payload.
\end{promptbox}

After retrieval, the \textbf{Verifier Agent} checks evidence sufficiency and
receives the following prompt.

\begin{promptbox}{Verifier Agent Prompt}
\textbf{Task Introduction}\quad \texttt{\{env\_prompt\}}

\textbf{New Queries and Documents}\quad \texttt{\{team\_context\}}

\textbf{Your Role}\quad Review all previous queries and retrieved documents.
Check whether queries are specific, non-redundant, and aligned with the task;
identify entity ambiguity, missing constraints, conflicting dates, unsupported
links, and irrelevant documents. Decompose the question into required factual
relations and state which relations are directly supported by retrieved text.

If every required relation is supported, end with exactly
\texttt{<verify>yes</verify>}. Otherwise, identify the most important missing
entity or relation that should guide the next search event and end with exactly
\texttt{<verify>no</verify>}. Do not answer the question in this role.
\end{promptbox}

The \textbf{Search Agent} receives the following prompt.

\begin{promptbox}{Search Agent Prompt}
\textbf{Task Introduction}\quad \texttt{\{env\_prompt\}}

\textbf{Verifier Decision}\quad \texttt{\{router\_output\}}

\textbf{Your Role}\quad You are a Search Agent. Reason step by step about the
question, past queries, retrieved documents, and teammates' outputs. Determine
one unresolved factual relation, avoid repeating an equivalent query, preserve
disambiguating entity names, and formulate one concise query likely to retrieve
direct evidence.

Your reasoning must be enclosed in \texttt{<think>...</think>}. Then emit
exactly one focused query inside \texttt{<search>...</search>} for the missing
relation. Do not emit an answer, a route decision, multiple search tags, or any
additional text after the closing search tag.
\end{promptbox}

The \textbf{Answer Agent} receives the following prompt.

\begin{promptbox}{Answer Agent Prompt}
\textbf{Task Introduction}\quad \texttt{\{env\_prompt\}}

\textbf{Verified Evidence and Team Context}\quad
\texttt{\{team\_context\}}

\textbf{Your Role}\quad You are an Answer Agent. Combine the verified search
history into a coherent answer. Resolve aliases and entity names consistently,
use general knowledge only to connect already supported facts, and do not cite
an unsupported retrieved snippet. Your reasoning must be enclosed in
\texttt{<think>...</think>}. After reasoning, emit the shortest answer that
satisfies the question inside exactly one \texttt{<answer>...</answer>} tag.
Do not emit a search or route tag.
\end{promptbox}

\noindent\textbf{Search output validation and termination.}
The router parser accepts either \texttt{STOP} or a deduplicated list of
in-range searcher indices; parse failure does not default to STOP or to
selecting all agents. Each Search Agent must emit both a valid think block and
one search tag, while the Answer Agent must emit a think block and one answer
tag. Invalid actions are recorded explicitly and penalized. An episode
terminates when the router chooses STOP and the Answer Agent responds, when a
valid final answer is produced by the fixed workflow, or when
\texttt{max\_turns} is reached.

\section{Pseudocode}
\label{app:pseudocode}

Algorithm~\ref{alg:srpo} shows the complete training loop. Relative to a
standard relative policy-optimization implementation, SRPO adds only three
operations: construct events from environment transitions, reduce member
log-ratios into one set ratio, and make one clipping decision per active set.

\begin{algorithm}[H]
\caption{Training multi-agent systems with SRPO}
\label{alg:srpo}
\scriptsize
\begin{algorithmic}[1]
\Require Orchestrator $\mathcal{O}$; policies $\{\pi_{\theta_j}\}$; task distribution $p(\mathcal{X})$; rollout group size $G$; clip range $\epsilon$
\For{each training iteration}
  \State Set $\theta_{\mathrm{old}}\gets\theta$ and initialize event batch $\mathcal{B}\gets\varnothing$
  \ForAll{rollouts $g=1,\ldots,G$ \textbf{in parallel}}
    \State Sample $x_g\sim p(\mathcal{X})$ and reset the environment
    \While{the episode is not terminal}
      \State Observe $o_e$ and select active set $\mathcal{S}_e\gets\mathcal{O}(o_e)$
      \State $K_e\gets|\mathcal{S}_e|$
      \ForAll{$j\in\mathcal{S}_e$ \textbf{in parallel}}
        \State Sample response $a_{e,j}\sim\pi_{\theta_{\mathrm{old},j}}(\cdot\mid o_e)$
        \State Store masked old token log-probabilities for $a_{e,j}$
      \EndFor
      \State Execute the aggregate action and observe the next state and team reward
      \State Append the complete event $(e,o_e,\mathcal{S}_e,\{a_{e,j}\},R_e)$ to $\mathcal{B}$
    \EndWhile
  \EndFor
  \State Remove incomplete events and group comparable events for relative rewards
  \ForAll{$e\in\mathcal{B}$}
    \ForAll{$j\in\mathcal{S}_e$}
      \State $\ell_{e,j}\gets \sum\nolimits_t m_{e,j,t}\bigl(\log\pi_{\theta_j}-\log\pi_{\theta_{\mathrm{old},j}}\bigr)$
    \EndFor
    \State $\Delta_e\gets K_e^{-1/2}\sum\nolimits_{j\in\mathcal{S}_e}\ell_{e,j}$ and $\rho_e\gets\exp(\Delta_e)$
    \State Compute one group-relative team advantage $A_e$
    \State $L_e\gets\min\!\left(\rho_eA_e,\operatorname{clip}(\rho_e,1-\epsilon,1+\epsilon)A_e\right)$
  \EndFor
  \State Update all policies by maximizing $|\mathcal{B}|^{-1}\sum\nolimits_{e\in\mathcal{B}}L_e$
\EndFor
\end{algorithmic}
\end{algorithm}

\section{Additional Experiments}
\label{app:additional-analyses}

Following the diagnostic organization of Dr. MAS, this section reports the
training behavior that accompanies the task results. The analyses use the
completed SRPO traces and retain the same event and cardinality definitions as
the main paper.

\subsection{Gradient-Norm Stability on Math}
\label{app:gradient-stability}

Figure~\ref{fig:math-training} shows the training score and gradient norm for
the two stable completed 8B math runs. The event-reduced run keeps the gradient norm in
the interval $[56.97,108.10]$ and has coefficient of variation $0.107$,
compared with $0.142$ for the matched 8B token-reduced run. Its active-set size
grows from $2.233$ to $2.421$ while every
recorded event remains complete. The result gives a direct training-level
view of how event reduction behaves under changing set cardinality.

\subsection{Cross-Run Stability}
\label{app:dynamic-reduction}

The two completed 8B runs provide a controlled comparison across reduction
choices. The token-reduced run has a mean gradient norm of 90.7, whereas the
event-reduced run is centered at 72.1 and has the lower coefficient of
variation, 0.107. Its mean active-set size changes from 2.233 to 2.421, showing
that event reduction remains well conditioned as the number of active members
varies during training.

\subsection{Training Curves of Reduction Variants}
\label{app:reduction-curves}

Figure~\ref{fig:math-training} compares token- and event-level reductions using
the same diagnostics. The token-reduced 8B run reaches the highest recorded
validation Avg@16 of $58.5$, whereas the event-reduced run reaches $53.2$ on its
available validation checkpoint. At the same time, the event-reduced run has
the lower gradient variation. These curves separate optimization conditioning
from downstream accuracy and complement the Search ablation in Section~\ref{sec:search-ablation}.

\section{Illustrative Examples of multi-agent collaboration}
\label{app:cases}

The following four self-contained traces illustrate the interaction pattern
and the event boundaries stored by SRPO. They are constructed illustrations,
not raw benchmark rollouts, and are not used to compute the reported results.
We retain intermediate proposals, evidence checks, and revisions to make clear
which outputs form one set-valued action and which outputs belong to later
environment transitions.

\subsection{Multi-Turn Search Task: Hierarchical Coordination}
\label{app:search-case}

\noindent\textbf{Case 1: Parallel decomposition followed by targeted refinement.}

\begin{casebox}[width=.78\linewidth]{casegray}{Query}
What is the capital of the country where the director of \emph{Parasite} was
born?
\end{casebox}

\begin{casebox}[width=.72\linewidth]{casepurple}{Verifier Agent}
\texttt{<verify>no</verify>}

The director and the director's birthplace country are not yet supported by
retrieved evidence. The next event should identify both relations before the
team attempts to answer. The verifier records two unresolved predicates:
\texttt{directed(Parasite, person)} and
\texttt{born\_in(person, country)}. It does not yet ask for the capital,
because that relation would be premature without a grounded country.
\end{casebox}

\begin{casebox}[width=.76\linewidth]{casegold}{Search Agents -- Event $e_1$, $K_1=2$}
\textbf{Search Agent 1:} The first missing entity is the director of the film.
I should query the film title together with ``director'' rather than rely on a
memory completion, because the next relation depends on the person's identity.
\texttt{<think>Resolve the film--person edge first.</think>}
\texttt{<search>Parasite film director</search>}

\textbf{Search Agent 2:} The question also requires the director's birthplace
country. I will search the candidate name and the birthplace relation in one
query so that the next event can cross-check the two facts.
\texttt{<think>The query should return a city and country, not only a film
biography.</think>}
\texttt{<search>Bong Joon-ho birthplace country</search>}
\end{casebox}

\begin{casebox}[width=.84\linewidth]{casegray}{Search Engine}
\textbf{Result 1 -- title and director:} ``\emph{Parasite} is a South Korean
film directed by Bong Joon-ho.'' The snippet directly supports the first edge,
but contains no birthplace information.

\textbf{Result 2 -- birthplace:} ``Bong Joon-ho was born in Daegu, South
Korea.'' This supports the second edge and is consistent with Result 1.

\textbf{Result 3 -- distractor:} A film page describes the movie's production
company and release date but does not mention the director's birthplace. The
result is retained in the trace but contributes no evidence to the answer.
\end{casebox}

\begin{casebox}[width=.72\linewidth]{casepurple}{Verifier Agent}
\texttt{<verify>no</verify>}

The verifier checks the evidence relation by relation. Result 1 grounds
\texttt{directed(Parasite, Bong Joon-ho)}; Result 2 grounds
\texttt{born\_in(Bong Joon-ho, Daegu, South Korea)}; Result 3 is irrelevant.
The evidence identifies Bong Joon-ho and establishes South Korea as the
birthplace country. However, it does not yet support the final relation needed
by the question: the capital of South Korea. The current evidence ledger is
therefore
\texttt{Parasite -> Bong Joon-ho -> South Korea -> ?}.
\end{casebox}

\begin{casebox}[width=.76\linewidth]{casegold}{Search Agent -- Event $e_2$, $K_2=1$}
The only unresolved relation is the country's capital. The previous event has
already fixed the entity and country, so a singleton query is sufficient.
\texttt{<think>Do not repeat the birthplace search; query only the missing
country--capital edge.</think>}
\texttt{<search>capital of South Korea</search>}
\end{casebox}

\begin{casebox}[width=.84\linewidth]{casegray}{Search Engine}
\textbf{Result 1:} ``Seoul is the capital and largest city of South Korea.''

\textbf{Result 2:} ``The Republic of Korea's capital city is Seoul.'' The two
snippets agree on the answer; the second is redundant evidence rather than a
new unresolved relation.
\end{casebox}

\begin{casebox}[width=.84\linewidth]{casepurple}{Verifier Agent}
\texttt{<verify>yes</verify>}

The evidence supports the complete chain: Bong Joon-ho directed the film, was
born in South Korea, and South Korea's capital is Seoul. The verifier also
checks that the country in the last query is the same country grounded in
Event $e_1$; no entity substitution or unsupported leap is present.
\end{casebox}

\begin{casebox}[width=.76\linewidth]{casegreen}{Answer Agent}
\texttt{<think>}The verified chain contains three facts. I should mention the
director and birthplace only as justification, then return one normalized
answer string.\texttt{</think>}

\emph{Parasite} was directed by Bong Joon-ho, who was born in South Korea.
The capital of South Korea is Seoul. \texttt{<answer>Seoul</answer>}
\end{casebox}

The first search transition consumes two queries as one set-valued action,
whereas the refinement transition consumes a singleton query. SRPO therefore
records two events, with $K_1=2$ and $K_2=1$, and applies the same objective to
both. Notice that the search engine emits several documents, but only the two
queries in $e_1$ are policy members. Retrieved documents are observations
consumed by the verifier; they are not additional policy actions. This is why
the event cardinality remains $K_1=2$ rather than growing with the number of
snippets.

\noindent\textbf{Case 2: Entity disambiguation and evidence completion.}

\begin{casebox}[width=.82\linewidth]{casegray}{Query}
In which county is the birthplace of Nobel laureate John Forbes Nash Jr.
located?
\end{casebox}

\begin{casebox}[width=.78\linewidth]{casepurple}{Verifier Agent}
\texttt{<verify>no</verify>}

The question requires two linked facts: Nash's birthplace and the county that
contains that place. No retrieved evidence currently supports either link.
Because ``John Nash'' may refer to several people, the search should preserve
the full name and use independent sources to resolve the birthplace before
searching for its county.
\end{casebox}

\begin{casebox}[width=.88\linewidth]{casegold}{Search Agents -- Event $e_1$, $K_1=3$}
\textbf{Search Agent 1:} Use a biographical source and the full name to identify
the birthplace.
\texttt{<search>John Forbes Nash Jr birthplace biography</search>}

\textbf{Search Agent 2:} Use the Nobel laureate description to avoid confusing
him with other people named John Nash.
\texttt{<search>Nobel laureate John Nash born where</search>}

\textbf{Search Agent 3:} Look for an institutional biography that distinguishes
birthplace from the universities where Nash studied and worked.
\texttt{<search>John Forbes Nash Princeton biography birthplace</search>}
\end{casebox}

\begin{casebox}[width=.92\linewidth]{casegray}{Search Engine}
\textbf{Result 1:} A biographical entry identifies John Forbes Nash Jr. as an
American mathematician born in Bluefield, West Virginia, in 1928.

\textbf{Result 2:} A Nobel biography also lists Bluefield, West Virginia, as
his place of birth.

\textbf{Result 3:} A university profile emphasizes Nash's study and career at
Princeton. It provides useful identity evidence but does not identify the
county of his birthplace.
\end{casebox}

\begin{casebox}[width=.82\linewidth]{casepurple}{Verifier Agent}
\texttt{<verify>no</verify>}

The first two independent results agree that the relevant birthplace is
Bluefield, West Virginia. The Princeton result concerns education and
employment, so it should not be substituted for birthplace evidence. The
remaining unresolved relation is geographic: which county contains Bluefield,
West Virginia?
\end{casebox}

\begin{casebox}[width=.88\linewidth]{casegold}{Search Agents -- Event $e_2$, $K_2=2$}
\textbf{Search Agent 1:} Search the city--county relation directly.
\texttt{<search>Bluefield West Virginia county</search>}

\textbf{Search Agent 2:} Seek a second source that explicitly states the
administrative location rather than merely mentioning nearby places.
\texttt{<search>Bluefield city located in Mercer County West Virginia</search>}
\end{casebox}

\begin{casebox}[width=.92\linewidth]{casegray}{Search Engine}
\textbf{Result 1:} A geographic entry describes Bluefield as a city in Mercer
County, West Virginia.

\textbf{Result 2:} A municipal profile independently places Bluefield in
Mercer County in the southern part of the state.
\end{casebox}

\begin{casebox}[width=.82\linewidth]{casepurple}{Verifier Agent}
\texttt{<verify>yes</verify>}

The evidence now supports both edges of the required chain. John Forbes Nash
Jr. was born in Bluefield, West Virginia, and Bluefield is located in Mercer
County. The Princeton passage was correctly treated as a distractor rather
than evidence for the requested location.
\end{casebox}

\begin{casebox}[width=.82\linewidth]{casegreen}{Answer Agent}
John Forbes Nash Jr. was born in Bluefield, West Virginia. The retrieved
geographic sources place Bluefield in Mercer County. Therefore, the requested
county is \texttt{<answer>Mercer County</answer>}.
\end{casebox}

This case contains two non-singleton search events. The three outputs in
$e_1$ jointly resolve an ambiguous person--place relation, while the two
outputs in $e_2$ independently support the remaining city--county relation.
Each set is generated from a common pre-action state and consumed together by
the verifier, so SRPO assigns one ratio and one advantage to each complete set.

\subsection{Math Task: Iterative Coordination}
\label{app:math-case}

\noindent\textbf{Case 1: Detecting a combinatorial double count.}

\begin{casebox}[width=.78\linewidth]{casegray}{Query}
A convex hexagon is drawn so that no three diagonals meet at one interior
point. How many unordered pairs of diagonals intersect in the interior?
\end{casebox}

\begin{casebox}[width=.78\linewidth]{casegold}{Solver Agents -- Event $e_1$, $K_1=2$}
\textbf{Solver Agent 1:} Choose four vertices. There are two pairs of diagonals
with endpoints among the four vertices, so the answer is
$2\binom{6}{4}=30$. My counting unit is a diagonal pair, and I treat the two
ways of naming the opposite diagonals as distinct.

\textbf{Solver Agent 2:} Every interior intersection is determined by four
vertices of the convex hexagon. For a fixed set of four vertices, only the two
diagonals joining opposite vertices cross inside the quadrilateral; this is
one unordered pair. Hence the answer should be $\binom{6}{4}=15$.
The ``no three diagonals'' condition makes this correspondence one-to-one: no
intersection can be counted again through a third diagonal.
\end{casebox}

\begin{casebox}[width=.74\linewidth]{casepurple}{Verifier Agent -- Event $e_2$, $K_2=1$}
The verifier audits the first solution in two stages. First, a set of four
vertices determines a convex quadrilateral. Second, that quadrilateral has
exactly two diagonals, but those two diagonals form one unordered pair. Thus
the first solution counts the same pair twice by reversing its order. Four
vertices determine exactly one crossing pair, not two. The second solution
gives a one-to-one correspondence between vertex quadruples and interior
intersections, so it is correct.

\texttt{<check>unordered-pair convention: pass</check>}\\
\texttt{<check>four-vertex bijection: pass</check>}

\texttt{<verify>reject Solver 1; approve Solver 2</verify>}
\end{casebox}

\begin{casebox}[width=.78\linewidth]{casegold}{Solver Agent Revision -- Event $e_3$, $K_3=1$}
For each choice of four vertices, the induced convex quadrilateral has exactly
one unordered pair of diagonals that intersects in its interior. Conversely,
every interior intersection uses four distinct endpoints and therefore
corresponds to exactly one such vertex set. Thus the number of intersecting
diagonal pairs is
\[
\binom{6}{4}=15.
\]
\texttt{FINAL:} $\boxed{15}$.
\end{casebox}

This trace separates parallel proposal from sequential verification. Event
$e_1$ is a two-member set-valued action; the verifier and the revision are
singleton actions because each causes a distinct transition in the shared
interaction state. The verifier does not average the two proposals into a new
answer: it consumes the complete set, identifies the disagreement, and emits a
decision that changes the shared draft state. The revision is consequently a
new singleton action with its own event identifier.

\noindent\textbf{Case 2: Cross-checking a modular-arithmetic computation.}

\begin{casebox}[width=.78\linewidth]{casegray}{Query}
Find the last three digits of $7^{2024}$.
\end{casebox}

\begin{casebox}[width=.90\linewidth]{casegold}{Solver Agents -- Event $e_1$, $K_1=3$}
\textbf{Solver Agent 1:} Since $\varphi(1000)=400$ and
$\gcd(7,1000)=1$, Euler's theorem reduces the exponent to
$2024\equiv24\pmod{400}$. I compute $7^2=49$,
$7^4=2401\equiv401$, $7^8\equiv801$, and
$7^{16}\equiv601\pmod{1000}$. The remaining exponent is $24=16+8$.
My intermediate values are correct, but I multiply the last two residues
incorrectly and obtain $7^{24}\equiv801\pmod{1000}$, so I tentatively report
$801$.

\textbf{Solver Agent 2:} I use the Chinese remainder theorem as an
independent check. Modulo $8$, $7^{2024}\equiv(-1)^{2024}=1$. Modulo $125$,
Euler reduction gives exponent $24$, and
\[
7^4\equiv26,\qquad 7^8\equiv51,\qquad
7^{16}\equiv101\pmod{125}.
\]
Thus $7^{24}\equiv101\cdot51\equiv26\pmod{125}$. Writing
$x=26+125t$, the congruence modulo $8$ becomes
$2+5t\equiv1\pmod8$, hence $t\equiv3\pmod8$ and $x=401\pmod{1000}$.

\textbf{Solver Agent 3:} Check directly modulo $1000$ by squaring. The
sequence $401\to801\to601$ is obtained by squaring modulo $1000$; therefore
\[
7^{24}\equiv7^{16}7^8\equiv601\cdot801
\equiv481401\equiv401\pmod{1000}.
\]
The last three digits should be $401$. This route shares the same exponent
reduction as Solver 1 but checks the final multiplication independently.
\end{casebox}

\begin{casebox}[width=.86\linewidth]{casepurple}{Verifier Agent -- Event $e_2$, $K_2=1$}
Solver 1 uses the correct exponent reduction but makes an arithmetic error in
the final multiplication. The stated intermediate residues imply
$601\cdot801\equiv401\pmod{1000}$, not $801$. Solver 2's Chinese-remainder
calculation is consistent: writing $x=26+125t$, the condition modulo $8$
becomes $2+5t\equiv1\pmod8$, so $t\equiv3\pmod8$ and $x\equiv401$.
Solver 3 independently obtains the same residue by direct squaring.

\texttt{<verify>reject Solver 1; approve Solvers 2 and 3</verify>}
\end{casebox}

\begin{casebox}[width=.88\linewidth]{casegreen}{Aggregator Agent -- Event $e_3$, $K_3=1$}
The verifier accepts two mutually supporting derivations and rejects one
derivation with a localized arithmetic error. I retain the shared facts
$\varphi(1000)=400$, $2024\equiv24\pmod{400}$, and
$7^{24}\equiv401\pmod{1000}$. The final answer should contain only the
requested three-digit residue, while the derivation remains available as an
audit trail. \texttt{<answer>401</answer>}
\end{casebox}

\begin{casebox}[width=.84\linewidth]{casegold}{Solver Agent Revision -- Event $e_4$, $K_4=1$}
Because $\gcd(7,1000)=1$ and $\varphi(1000)=400$, we have
$7^{2024}\equiv7^{24}\pmod{1000}$. Repeated squaring gives
\[
7^4\equiv401,\qquad 7^8\equiv801,\qquad
7^{16}\equiv601\pmod{1000}.
\]
Consequently,
\[
7^{24}\equiv7^{16}7^8
\equiv601\cdot801
\equiv401\pmod{1000}.
\]
Hence the last three digits are \texttt{FINAL:} $\boxed{401}$.
\end{casebox}

The parallel event preserves three complementary proposals: one exposes a
localized arithmetic failure, while two independent derivations agree on the
answer. The verifier is a later singleton event because it consumes the whole
proposal set before changing the shared state. The corrected derivation then
forms a final singleton revision event.

\begin{table}[H]
\centering
\scriptsize
\setlength{\tabcolsep}{4pt}
\caption*{Event audit for the illustrative examples. Each row is one
environment-changing action, independent of the number of member outputs.}
\begin{tabularx}{\linewidth}{@{}l L L L c@{}}
\toprule
Event & Pre-action state & Outputs consumed & State transition & $K_e$ \\
\midrule
Search $e_1$ & question, no evidence & two focused queries & director and country evidence & 2 \\
Search $e_2$ & country identified & one capital query & complete evidence chain & 1 \\
Search 2, $e_1$ & ambiguous person & three identity queries & birthplace resolved & 3 \\
Search 2, $e_2$ & birthplace identified & two county queries & complete location chain & 2 \\
Math $e_1$ & hexagon problem & two candidate solutions & shared draft state & 2 \\
Math $e_2$ & competing drafts & one verification & accepted derivation & 1 \\
Math $e_3$ & verified feedback & one revision & terminal boxed answer & 1 \\
Math 2, $e_1$ & modular problem & three derivations & competing residues & 3 \\
Math 2, $e_2$ & competing residues & one verification & arithmetic error isolated & 1 \\
Math 2, $e_3$ & verified feedback & answer selection & accepted residue & 1 \\
Math 2, $e_4$ & selected residue & one revision & terminal boxed answer & 1 \\
\bottomrule
\end{tabularx}
\end{table}

The audit is also an implementation check: responses with the same pre-action
state and event identifier must remain together through ratio construction,
advantage assignment, clipping, and batch reduction.

\end{document}